\documentclass[twoside,11pt]{article}

\newcommand{\eg}{e.\,g.,\ }

\usepackage{amsfonts}
\usepackage{amssymb,amsmath,bm,bbm}
\usepackage{subfig}
\usepackage{graphicx}
\usepackage{enumitem}
\usepackage[activate]{microtype}

\usepackage[abbrvbib,preprint]{jmlr2e}

\jmlrheading{volume}{2020}{pages}{date submitted}{date published}{paper id}{Felix Weninger, Yue Zhang, Rosalind W. Picard}

\ShortHeadings{openXDATA: A Tool for Multi-Target Data Generation and Missing Label Completion}{Weninger, Zhang and Picard}
\firstpageno{1}

\begin{document}

\title{openXDATA: A Tool for Multi-Target Data Generation and Missing Label Completion}

\author{\name Felix Weninger \email felix@weninger.de \\
      \addr Nuance Communications \\
      Burlington, MA 01803, USA
      \AND
      \name Yue Zhang \email yuefw@mit.edu
      \\
      \name Rosalind W.\ Picard \email picard@mit.edu \\
      \addr Affective Computing Group, MIT Media Lab\\
      Massachusetts Institute of Technology\\
      Cambridge, MA 02139, USA}

\editor{}

\maketitle

\begin{abstract}
A common problem in machine learning is to deal with datasets with disjoint label spaces and missing labels. 
In this work, we introduce the openXDATA tool that completes the missing labels in partially labelled or unlabelled datasets in order to generate multi-target data with labels in the joint label space of the datasets. 
To this end, we designed and implemented the cross-data label completion (CDLC) algorithm that uses a multi-task shared-hidden-layer DNN to iteratively complete the sparse label matrix of the instances from the different datasets. 
We apply the new tool to estimate labels across four emotion datasets: one labeled with discrete emotion categories (\eg happy, sad, angry), one labeled with continuous values along arousal and valence dimensions, one with both kinds of labels, and one unlabeled. 
Testing with drop-out of true labels, we show the ability to estimate both categories and continuous labels for all of the datasets, at rates that approached the ground truth values. 
openXDATA is available under the GNU General Public License from \url{https://github.com/fweninger/openXDATA}.
\end{abstract}

\begin{keywords}
  cross-data label completion, pseudo-labeling, multi-target, multi-task
\end{keywords}

\section{Introduction}

Multi-target and multi-task learning are related, yet different concepts in machine learning. 
Multi-target learning deals with learning from examples that have multiple target attributes. 
Multi-label classification (e.g.\ tagging) can been seen as a special case of multi-target learning where the labels are binary \citep{Zhang2013-ARO}. 
Multi-task learning is an approach to inductive transfer by learning tasks in parallel, where a ``task'' refers to a target attribute \citep{Caruana1997-ML}. 
In contrast to multi-target learning, the examples do not necessarily share the same label space, that is to say, they can be taken from different single- or multi-target datasets. 
There exist a number of tools and libraries for multi-label and multi-target learning, \eg \textsc{Meka} \citep{Read2016-MAM}, \textsc{Mulan} \citep{Tsoumakas2011-MAJ} and \textsc{Scikit-multilearn} \citep{Szymanski2019-SMA}. To employ these tools or to run statistical analysis on multiple target attributes, one first needs to acquire multi-target data.  
Due to the bottleneck of manual data annotation, many existing datasets are labelled along one or a few target dimensions, which might also be attributable to the traditional supervised learning paradigm. 
For example, emotion datasets usually have different labeling schemes due to the diversity of emotional concepts \citep{Zhang20-HAR}. 
Another challenge is handling data with missing labels that can happen for various reasons in the data collection process. 
Moreover, there is a huge potential to improve model performance and the quality of predicted labels by leveraging completely unlabelled data. 

In this work, we introduce the openXDATA tool to generate multi-target data from any input datasets within the same feature space for any modality. 
There is no constraint on the label space of the specific datasets, which can be single-target, multi-target, incomplete with missing labels, or completely unlabelled; nor on the labelling scheme (nominal, numeric). 
The objective of the openXDATA tool is to create a holistic database that is labelled along all the target dimensions of the input datasets. The key idea is to extend the cross-task labelling (CTL) algorithm \citep{Zhang2016-DAS} to multi-task models and multi-target datasets.

\section{Algorithm}

\begin{figure}[t]
    \centering\hspace*{\fill}
    \subfloat[Label matrix $\bf Y$]{
    \begin{minipage}[b][7.5cm][b]{0.32\linewidth}
    \includegraphics[width=\textwidth]{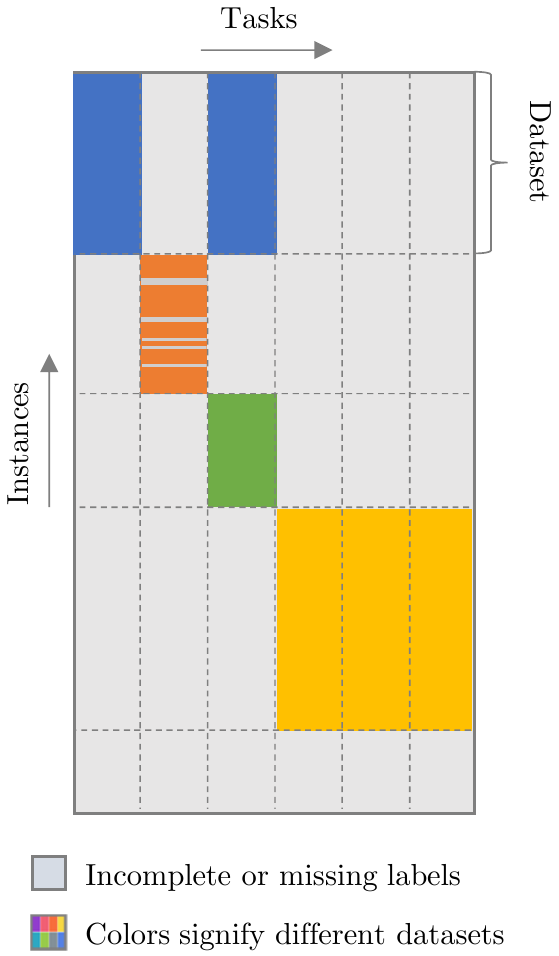}
    \end{minipage} \hspace{1.2cm}
    \label{fig:xdata}}
    \subfloat[Pseudocode]{
    \begin{minipage}[b][7.5cm][b]{0.5\linewidth}
    \begin{flushleft}
    \small
    	{\rule{\textwidth}{.4pt}}\\
    	{\bf Algorithm}: {\em Cross-Data Label Completion}\\
    	{\bf Input}: original data sets $\mathcal{D} = \{ ( {\bf x}_i^{(d)}, {\bf y}_i^{(d)} ) \}$, $1 \leq d \leq D$\\
    	{\bf Output}: completed multi-target data set $\{ ( {\bf x}_i, {\bf y}_i = (y_{i,1}, \dots, y_{i,M})^\intercal \}$ \\
    	{\bf Initialisation}: {\bf X}, {\bf Y} = MultiTargetDataset($\mathcal{D}$) \\
    	{\bf Do}: \\
        \quad $\mathcal{L} := \{ ({\bf x}_i, {\bf y}_i) \mid \exists m: y_{i,m} \neq \bot \}$ \\
        \quad $\mathcal{U} := \{ ({\bf x}_i, {\bf y}_i) \mid \exists m: y_{i,m} = \bot \}$ \\
    	\quad $h := \text{Train}(\mathcal{L})$ \\
    	\quad $\hat{Y} := \text{Predict}(h, \mathcal{U}) $\\
    	\quad {\bf For} $m = 1, \dots, M$: \\
    	\quad \quad $\mathcal{I}_m := \text{Select}(\hat{Y}_m, C_m)$ \quad // by highest $C_m$\\
    	\quad \quad {\bf For} $i \in \mathcal{I}_m$: $y_{i,m} := \hat{y}_{i,m}$ \\
    	{\bf While} $\mathcal{U}^{(l)} \neq \emptyset$ \\
    	{\rule{\textwidth}{.4pt}}
    \end{flushleft}
    \end{minipage}
    \label{fig:algorithm}} 
    \caption{Multi-target data generation and missing label completion using the Cross-Data Label Completion (CDLC) algorithm with multi-task shared-hidden-layer DNN.}
    \vspace{-0.5cm}
\end{figure}

\figurename~\ref{fig:xdata} illustrates our problem space. 
Given a set of unlabelled or partially labelled datasets, the goal of the CDLC algorithm is to complete the missing labels in the label matrix in order to generate multi-target data. 
The input datasets can be single-target (orange, green) or multi-target (blue, yellow), they may have overlapping tasks (blue and green), missing labels (orange), or they can also be completely unlabelled (last row in grey). 

From the input datasets $d = 1, \dots, D$, we determine the set of tasks $\mathcal{M} = \{1, \dots, M\}$ as the union of the tasks associated with the datasets. 
Joining the instances $i = 1, \dots, N$ of all the datasets, we build a feature matrix ${\bf X} = [{\bf x}_1^\intercal, \cdots, {\bf x}_N^\intercal]^\intercal$
and a label matrix ${\bf Y} = [{\bf y}_1^\intercal, \cdots, {\bf y}_N^\intercal]^\intercal$.
There, ${\bf y}_i \in \mathcal{Q}_1 \times \cdots \times \mathcal{Q}_M$ is the label vector for instance $i$, and $\mathcal{Q}_m$ is the label set for task $m$. 
Note that $\mathcal{Q}_m$ also contains the symbol for missing labels $\bot$.
For example, let us assume that $\mathcal{Q}_1 = \{0,1,\bot\}$ is binary-class, $\mathcal{Q}_2 = \{0,1,2,3,\bot\}$ is multi-class, and $\mathcal{Q}_3 = \mathbb{R} \cup \{\bot\}$ contains real-valued labels. 
As shown in \figurename~\ref{fig:xdata}, the label matrix ${\bf Y}$ is generally sparse, containing many undefined labels $\bot$.

\figurename~\ref{fig:algorithm} shows the pseudocode of the CDLC algorithm, which extends the previous CTL algorithm based on pseudo-labeling \citep{Lee2013-PLT} and self-training \citep{Rosenberg2005-SSS}. 
The CDLC algorithm trains a multi-task model on the instances with at least one label ($\mathcal{L}$), and makes predictions for the instances with missing labels ($\mathcal{U}$). 
For each task $m$, the predictions with the highest model confidence $C_m$ are added to the label matrix. 
The model is then retrained and the algorithm continues iteratively until the label matrix is completed or a stopping criterion is fulfilled. 
Thus, the grey part in \figurename~\ref{fig:xdata} receives generated labels, learned both within and across all of the different datasets and tasks.

\section{Design and Implementation}

The code is written in Python and is organized in the following modules:

\begin{itemize}[itemsep=1pt,topsep=1pt,leftmargin=*]
    \item {\em MultiTargetDataset}: constructs the feature and label matrices; reads and writes data files in the Attribute Relation File Format (ARFF) \citep{Hall2009-TWD}
    \item {\em Model}: implements the multi-task shared-hidden-layer (MT-SHL) algorithm for training deep neural networks (DNN) \citep{Huang2013-CLK}; computes predictions and confidence measures \citep{Gal2016-UID}
    \item {\em Options}: provides dataset specifications (file names and number of target variables in each file) and various configuration variables (number of hidden layers, training epochs, learning rate, etc.) to be set by the user
    \item {\em Trainer}: implements the CDLC algorithm according to \figurename~\ref{fig:algorithm}, optionally providing performance measurements on an evaluation set
\end{itemize}
The output $\hat{\bf y}_i = [ \hat{\bf y}_{i,1}; \hat{\bf y}_{i,2}; \dots; \hat{\bf y}_{i,M} ]$ of the MT-SHL-DNN for an input vector ${\bf x}_i$ is composed of sub-vectors for each task $m=1, \dots, M$.
Each $\hat{\bf y}_{i,m}$, $m = 1, \dots, M$ corresponds to a task-specific transformation of the topmost shared hidden layer activation. 
This transformation consists of (optional) hidden layers and an output layer.
The number of shared hidden layers can be set by the user.
The following MT loss function is minimized: 
\begin{equation}
    J^{\text{MT}} = \sum_i \sum_m \mathbbm{1}( y_{i,m} \neq \bot ) J (\hat{\bf y}_{i,m}, y_{i,m}) .
    \label{eq:l_mt}
\end{equation}
For the output layer activation functions and task-specific loss functions $J$, we use the sigmoid function with cross-entropy loss in binary classification, the softmax function with cross-entropy in multi-class classification, and the linear activation function with mean squared error loss in regression.

For obtaining the confidence $C_m$ in \figurename~\ref{fig:algorithm}, we perform a fixed number of forward passes with different dropout masks. In case of (binary or multi-class) classification, we compute the Shannon entropy of the average output(s), and for regression, we take the variance of the output, following \citet{Gal2016-UID}.

\section{Experiments}

\begin{figure}[t]
    \vspace{-.5cm}
    \centering
    \subfloat[Visualization of generated data set]{
    \raisebox{-.5\height}{
    \includegraphics[width=0.5\textwidth]{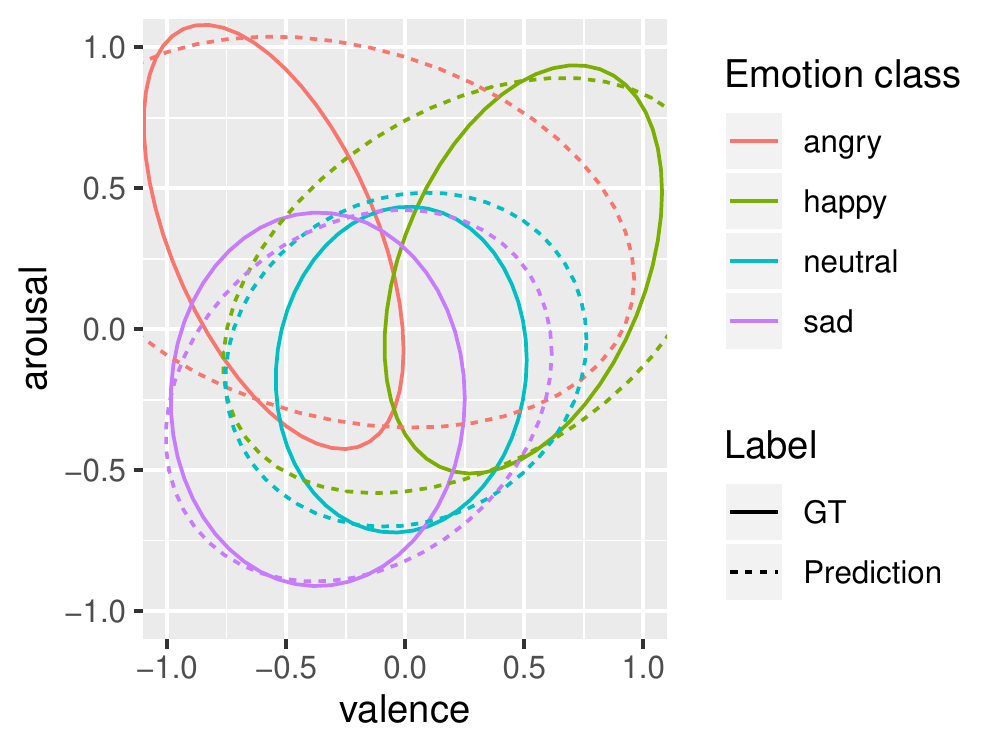}
    }
    \label{fig:labels}} 
    \subfloat[Test set performance]{
    \small
    \begin{tabular}{c|c|c|c}
         \# & UAR (E) & CC (A) & CC (V) \\ \hline
         0 & .562 & .701 & .342 \\ 
         1000 & .562 & .704 &	.380 \\ 
         2000 & .563 & .703 & .410 \\
         3000 & .569 & .708 & .414 \\
         4000 & .576 & .715 & .427 \\ \hline
         \em GT & \em .581 & \em .735 & \em	.453 \\
    \end{tabular}
    \label{fig:results}}
    \caption{Evaluation of the toolkit on multi-target emotion recognition with missing labels.
    \#: number of cross-labeled instances. UAR: unweighted average recall (E: emotion classes). CC: correlation coefficient (A: arousal, V: valence). 
    GT: using ground-truth labels for all instances (upper bound on performance). 
    }
    \vspace{-0.5cm}
\end{figure}

To validate the functionality of the openXDATA tool, we use it to generate multi-target emotion data, which are particularly rare given the prevalence of small datasets with task-specific labeling schemes (\eg categorical, dimensional). 
For the experiments, we use the IEMOCAP speech corpus \citep{Busso2008-IIE} to simulate exemplary datasets with disjoint label spaces and missing labels. 
The first recording session is labeled in four emotion categories (angry, happy, sad, neutral), as well as arousal/\,valence  on a continuous scale. For the second one we only use the emotion class labels, and for the third only the arousal/\,valence labels. The fourth one is treated as completely unlabeled. These four sessions form the cross-labeling set while the fifth one serves as the test set. In addition, we simulate missing labels by dropping 75\,\% of the labels from the cross-labeling set. 
The feature files and detailed parameter settings are included with the toolkit.

As a qualitative evaluation of the labels obtained by CDLC, we plot the distribution of arousal and valence by the emotion categories (GT: ground-truth and predicted) on the cross-labeled instances in \figurename~\ref{fig:labels}.
Both GT and predicted emotions show the same trend in light of the circumplex model of affect.
The deviation can be attributed to the well-known difficulty of recognizing valence from speech.
Furthermore, \figurename~\ref{fig:results} shows the performance on the test set after each iteration of the CDLC algorithm (1000 instances cross-labeled per iteration and task). It can be seen that the performance gradually approaches the one obtained by using the corresponding GT labels.

\section{Conclusion}

In this paper, we introduced the openXDATA toolkit and demonstrated its functionality. Due to its flexibility regarding input features, missing labels and classification/\,regression tasks, it has the potential to be applied to many different machine learning tasks.

\newpage
\bibliography{jmlr}

\begin{thebibliography}{13}
\providecommand{\natexlab}[1]{#1}
\providecommand{\url}[1]{\texttt{#1}}
\expandafter\ifx\csname urlstyle\endcsname\relax
  \providecommand{\doi}[1]{doi: #1}\else
  \providecommand{\doi}{doi: \begingroup \urlstyle{rm}\Url}\fi

\bibitem[Busso et~al.(2008)Busso, Bulut, Lee, Kazemzadeh, Mower, Kim, Chang,
  Lee, and Narayanan]{Busso2008-IIE}
C.~Busso, M.~Bulut, C.-C. Lee, A.~Kazemzadeh, E.~Mower, S.~Kim, J.~Chang,
  S.~Lee, and S.~Narayanan.
\newblock {IEMOCAP}: Interactive emotional dyadic motion capture database.
\newblock \emph{Language Resources and Evaluation}, 42\penalty0 (4):\penalty0
  335--359, 2008.

\bibitem[Caruana(1997)]{Caruana1997-ML}
R.~Caruana.
\newblock Multitask learning.
\newblock \emph{Machine Learning}, 28\penalty0 (1):\penalty0 41--75, 1997.

\bibitem[Gal(2016)]{Gal2016-UID}
Y.~Gal.
\newblock \emph{Uncertainty in Deep Learning}.
\newblock PhD thesis, University of Cambridge, Cambridge, UK, 2016.

\bibitem[Hall et~al.(2009)Hall, Frank, Holmes, Pfahringer, Reutemann, and
  Witten]{Hall2009-TWD}
M.~Hall, E.~Frank, G.~Holmes, B.~Pfahringer, P.~Reutemann, and I.~H. Witten.
\newblock The weka data mining software: An update.
\newblock \emph{ACM SIGKDD Explorations Newsletter}, 11\penalty0 (1):\penalty0
  10--18, 2009.

\bibitem[Huang et~al.(2013)Huang, Li, Yu, Deng, and Gong]{Huang2013-CLK}
J.-T. Huang, J.~Li, D.~Yu, L.~Deng, and Y.~Gong.
\newblock Cross-language knowledge transfer using multilingual deep neural
  network with shared hidden layers.
\newblock In \emph{Proc.\ of 38th IEEE International Conference on Acoustics,
  Speech, and Signal Processing (ICASSP)}, pages 7304--7308, Vancouver, Canada,
  2013. IEEE.

\bibitem[Lee(2013)]{Lee2013-PLT}
D.-H. Lee.
\newblock Pseudo-label: The simple and efficient semi-supervised learning
  method for deep neural networks.
\newblock In \emph{Proc.\ of 30th International Conference on Machine Learning
  (ICML), Workshop on challenges in representation learning}, volume~3, page~2.
  IMLS, 2013.

\bibitem[Read et~al.(2016)Read, Reutemann, Pfahringer, and
  Holmes]{Read2016-MAM}
J.~Read, P.~Reutemann, B.~Pfahringer, and G.~Holmes.
\newblock Meka: A multi-label/multi-target extension to weka.
\newblock \emph{Journal of Machine Learning Research}, 17:\penalty0 667--671,
  2016.

\bibitem[Rosenberg et~al.(2005)Rosenberg, Hebert, and
  Schneiderman]{Rosenberg2005-SSS}
C.~Rosenberg, M.~Hebert, and H.~Schneiderman.
\newblock Semi-supervised self-training of object detection models.
\newblock In \emph{Proc.\ of IEEE Workshop on Motion and Video Computing},
  pages 29--36, Breckenridge, CO, 2005. IEEE.

\bibitem[Szyma{{\'n}}ski and Kajdanowicz(2019)]{Szymanski2019-SMA}
P.~Szyma{{\'n}}ski and T.~Kajdanowicz.
\newblock scikit-multilearn: A python library for multi-label classification.
\newblock \emph{Journal of Machine Learning Research}, 20\penalty0
  (6):\penalty0 1--22, 2019.

\bibitem[Tsoumakas et~al.(2011)Tsoumakas, Spyromitros-Xioufis, Vilcek, and
  Vlahavas]{Tsoumakas2011-MAJ}
G.~Tsoumakas, E.~Spyromitros-Xioufis, J.~Vilcek, and I.~Vlahavas.
\newblock Mulan: A java library for multi-label learning.
\newblock \emph{Journal of Machine Learning Research}, 12:\penalty0 2411--2414,
  2011.

\bibitem[Zhang and Zhou(2013)]{Zhang2013-ARO}
M.-L. Zhang and Z.-H. Zhou.
\newblock A review on multi-label learning algorithms.
\newblock \emph{IEEE Transactions On Knowledge and Data Engineering},
  26\penalty0 (8):\penalty0 1819--1837, 2013.

\bibitem[Zhang et~al.(2016)Zhang, Weninger, Ren, and Schuller]{Zhang2016-DAS}
Y.~Zhang, F.~Weninger, Z.~Ren, and B.~Schuller.
\newblock Sincerity and deception in speech: Two sides of the same coin? {A}
  transfer- and multi-task learning perspective.
\newblock In \emph{Proc.\ 17th Annual Conference of the International Speech
  Communication Association (INTERSPEECH)}, pages 2041--2045, San Francisco,
  CA, 2016. ISCA.

\bibitem[Zhang et~al.(2020)Zhang, Weninger, Schuller, and Picard]{Zhang20-HAR}
Y.~Zhang, F.~Weninger, B.~Schuller, and R.~Picard.
\newblock Holistic affect recognition using {PaNDA}: Paralinguistic non-metric
  dimensional analysis.
\newblock \emph{IEEE Transactions on Affective Computing}, 2020.
\newblock to appear.

\end{thebibliography}

\end{document}